\documentclass[letterpaper, 10 pt, conference]{ieeeconf}  
\IEEEoverridecommandlockouts                              

\usepackage{graphics}   
\usepackage{times}     
\usepackage{amsmath}   
\usepackage{amssymb}   
\usepackage{graphicx}
\usepackage{tabularx}
\usepackage{booktabs}
\usepackage{multirow}
\usepackage{xcolor} 
\usepackage[separate-uncertainty=true,number-unit-product=\ ]{siunitx}
\usepackage{subcaption}
\usepackage{caption}
\usepackage[hidelinks]{hyperref} 
\usepackage{makecell}

\definecolor{SafetyColor}{rgb}{0.6, 0.3, 0.0}
\definecolor{EfficiencyColor}{rgb}{0.0, 0.6, 0.0}
\definecolor{AggressivenessColor}{rgb}{0.0, 0.0, 0.6}
\definecolor{ComfortColor}{rgb}{0.541, 0.203, 0.619}
\definecolor{SpeedColor}{rgb}{1.0, 0.647, 0.0}

\usepackage{colortbl} 
\definecolor{lightgray}{gray}{0.94}
\definecolor{ddgray}{gray}{0.83}
\definecolor{darkgray}{gray}{0.87}
\definecolor{lgray}{gray}{0.99}

\newcolumntype{Y}{>{\centering\arraybackslash}p{1.575cm}}

\def\eqref#1{Eq.~(\ref{#1})}

\newcommand\etal{\emph{et al. }}

\newenvironment{rqenum}{
	
	\begin{enumerate}
	}{
	\end{enumerate}
}

\title{\LARGE \bf Multi-Objective Reinforcement Learning for Adaptable Personalized Autonomous Driving}

\author{Hendrik Surmann \and Jorge de Heuvel \and Maren Bennewitz
  \thanks{
    All authors are with the University of Bonn, Germany. 
   M. Bennewitz and J. de Heuvel are additionally with the Lamarr Institute for Machine Learning and Artificial Intelligence, and the Center for Robotics, Bonn, Germany. 
   This work has been partially funded by the Robotics Institute Germany, grant No. 16ME0999, and by the Federal Ministry of Education and Research of Germany and the state of North-Rhine Westphalia as part of the Lamarr Institute for Machine Learning and Artificial Intelligence, LAMARR22B.
   }
}

\begin{document}
\maketitle
\thispagestyle{empty} 
\pagestyle{empty}

\begin{abstract}
Human drivers exhibit individual preferences regarding driving style.
Adapting autonomous vehicles to these preferences is essential for user trust and satisfaction.
However, existing end-to-end driving approaches often rely on predefined driving styles or require continuous user feedback for adaptation, limiting their ability to support dynamic, context-dependent preferences.
We propose a novel approach using multi-objective reinforcement learning (MORL) with preference-driven optimization for end-to-end autonomous driving that enables runtime adaptation to driving style preferences.
Preferences are encoded as continuous weight vectors to modulate behavior along interpretable style objectives—including efficiency, comfort, speed, and aggressiveness—without requiring policy retraining.
Our single-policy agent integrates vision-based perception in complex mixed-traffic scenarios and is evaluated in diverse urban environments using the CARLA simulator. Experimental results demonstrate that the agent dynamically adapts its driving behavior according to changing preferences while maintaining performance in terms of collision avoidance and route completion.
Our code is available online on \url{https://github.com/HumanoidsBonn/morl_preference_driving}.
\end{abstract} 

\vspace*{-0.4em}
\section{Introduction}
\vspace*{-0.25em}
\label{sec:intro}

\begin{figure}[t]
	\centering
	\includegraphics[width=0.99\linewidth]{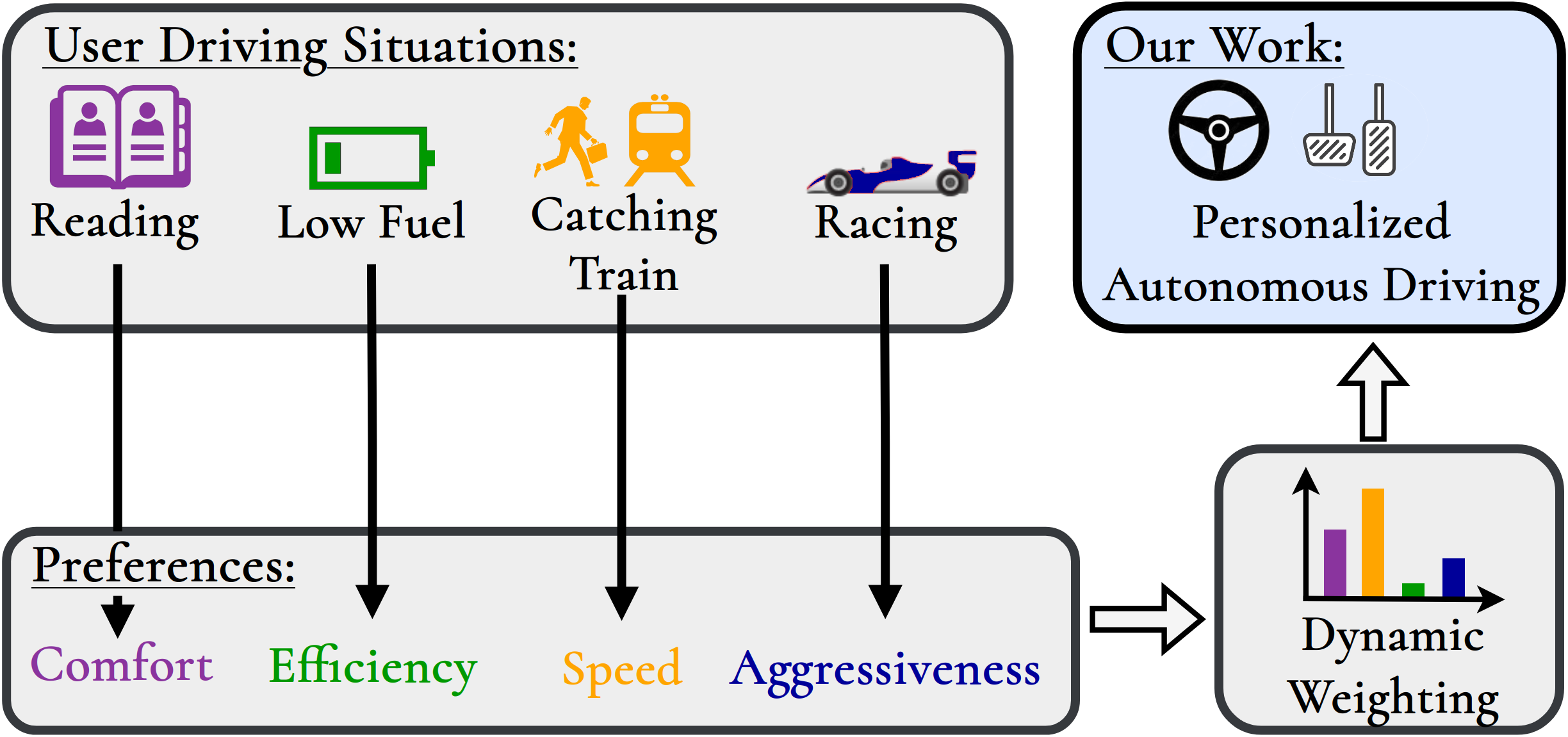}   
    \caption{
    Diverse driving situations necessitate distinct driving styles, motivating preference objectives such as comfort, efficiency, aggressiveness, and speed. 
    By providing our agent with user-defined priorities over these objectives, expressed via dynamic weightings, it learns to personalize autonomous driving behavior accordingly.
    }
	\label{fig:motivations}
\end{figure}

Recent advances in end-to-end autonomous driving (AD) have enabled reliable navigation in complex urban scenarios \cite{waymo, nvidiaWinner}, and autonomous vehicles are becoming increasingly integrated into real-world environments. To ensure user satisfaction, trust, and long-term acceptance \cite{motivation, Trust}, autonomous vehicles must be capable of aligning their behavior with individual user preferences.
Human drivers exhibit distinct and dynamic preferences regarding driving style that may vary with time, context, and task \cite{dynamic_prefs, User-DrivenAD}, see Figure~\ref{fig:motivations}.

However, existing vision-based end-to-end AD approaches typically rely on static preference configurations using lexicographic multi-objective reinforcement learning (MORL)~\cite{lexicoMorl, urban-morl, MultiMorl}, or require extensive human expert input \cite{PORF-DDPG}.
Others necessitate re-optimization whenever preferences change \cite{Personal_MORL}, or learn multiple policies to approximate preference trade-offs \cite{2objMorl}, all offering limited support for efficient and dynamic preference adaptation.

Data-driven methods that extract driving styles from demonstrations \cite{IRLsytles, frequent, Rosbach_2019} heavily rely on curated datasets that often lack the diversity needed to model fine-grained or context-dependent preferences and typically struggle to generalize to unseen tasks or driving conditions.
Furthermore, recent work exploring driving behavior modulation via natural language remains unsuitable for real-time use \cite{LLMstyleCarla} or is limited to highly abstract simulation settings \cite{wordsToWheel}.

In particular, personalizing driving behavior by incorporating dynamic user preferences into vision-based, end-to-end autonomous driving remains an open research challenge.

To address this, we propose a single-policy, preference-driven multi-objective reinforcement learning approach for end-to-end AD.
Preferences are encoded as a continuous weight vector and provided at runtime to modulate the agent's behavior along different driving styles, namely comfort, efficiency, speed, and aggressiveness. 
Our setup allows post-training adaptation to diverse and changing user preferences, without retraining or maintaining multiple policies.
 
Our approach builds on vision-based perception and operates in dense, mixed-traffic urban environments using the CARLA simulator.
To support generalization and flexible input configurations, we explore different feature extraction pipelines and multi-modal input modalities. 
Evaluation in diverse scenarios shows that the agent dynamically adapts its driving style to match varying preferences while maintaining robust navigation and collision avoidance. 

\begin{figure*}[t]
	\centering
	\includegraphics[width=0.99\linewidth]{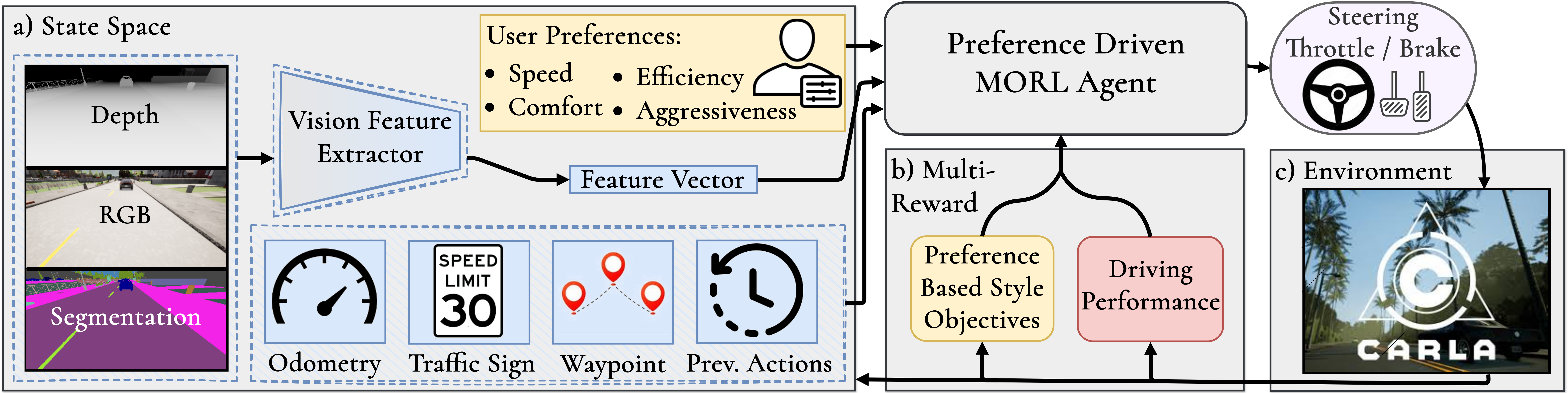}
    \caption[Overview of our approach]{
    Our approach enables personalized autonomous driving by integrating user preferences into a multi-objective reinforcement learning setup.   
    \textbf{a)} The agent's state space consists of multiple inputs, including camera data (RGB, depth, or segmentation), which are processed by a vision feature extractor (e.g., ResNet, MobileNet, or a custom CNN) to produce a low-dimensional feature vector. This vector, combined with user-defined preference weights and additional state inputs—including vehicle odometry, the current speed limit, navigation instructions to the next waypoint, and the agent’s previous three actions—serves as input to the MORL agent.
    \textbf{b)} The multi-dimensional reward consists of core driving performance and preference-based driving style objectives, forming a six-dimensional vector.
    \textbf{c)} The CARLA environment executes the agent’s continuous actions (steering and throttle/brake) and returns the next state and reward, enabling reinforcement learning.
    }
	\label{fig:approach}
\end{figure*}

The primary contributions of our work are threefold:
\begin{itemize} 
    \item We propose a vision-based, single-policy MORL approach for preference-reflecting, end-to-end autonomous driving.
 
    \item We design a reward vector that separates the core driving task from preference-based style objectives, enabling smooth transitions across the driving styles comfort, speed, aggressiveness and efficiency.
 
    \item We present an extensive evaluation with ablations on feature extractors and input modalities, demonstrating robust, preference-aware behavior in mixed-traffic urban scenarios using the CARLA simulator.
\end{itemize} 

\section{Related Work}
\vspace*{-0.05em}
\label{sec:related}

Personalization in autonomous driving is increasingly recognized as essential for enhancing user experience, trust, satisfaction, and long-term acceptance by aligning vehicle behavior with individual needs~\cite{User-DrivenAD, onboardLLM,  motivationReview}. 
Traditional approaches explored this alignment through control-based methods~\cite{occupant, controllerPersonal}. Li et al.~\cite{traditional} embed personalized constraints into a Model Predictive Control (MPC) framework extended with proportional–integral–derivative (PID) feedback to reflect individual driving styles via explicit control parameter tuning. However, such methods depend on handcrafted control rules and are often limited to specific scenarios, lacking the scalability of learning-based approaches.

As research shifts toward end-to-end autonomous driving frameworks~\cite{survey}, inverse reinforcement learning (IRL) methods address personalization by extracting reward functions from human demonstrations~\cite{IRLsytles, Rosbach_2019, IRL}, enabling imitation of naturalistic driving behavior. However, these approaches often suffer from limited generalization and rely on curated datasets, restricting adaptability to diverse, dynamic preferences~\cite{wordsToWheel}.

Chen~\textit{et al.}~\cite{PORF-DDPG} extend this direction with a human-in-the-loop RL approach that refines reward models through interactive supervision. While effective in capturing user-aligned behavior, the method requires frequent human feedback, limiting scalability.

Language-based approaches have recently emerged to modulate driving behavior via natural language commands~\cite{LLMstyleCarla, wordsToWheel, texttodrive}. While promising, these methods lack applicability in realistic real-time, vision-based driving environments, as they are either confined to abstract simulations or suffer from high latency. Cui et al.~\cite{onboardLLM} propose an onboard vision-language model (VLM) framework for near real-time personalized driving via natural language. However, their system relies on discrete, instruction-based feedback and is limited to narrow, predefined scenarios, lacking support for continuous preference modulation and complex urban environments, as addressed in our work.

Multi-objective reinforcement learning extends standard RL to vector-based reward functions with policies that learn to balance competing objectives, making it particularly well-suited for preference-driven end-to-end AD.
Li and Czarnecki~\cite{urban-morl} apply thresholded lexicographic Q-learning to urban driving, optimizing fixed objectives such as safety and comfort, each managed by a separate agent, but without vision-based inputs as addressed in our approach.
Hu et al.~\cite{MultiMorl} propose a MORL framework that balances speed and comfort in highway-like scenarios using a Unity-based simulator. However, their setup is limited by simplified dynamics utilizing discrete high-level actions, as well as by only two objectives and fixed preferences. In contrast, our approach supports dynamic preference integration in complex urban environments with a broader range of personalized driving styles.
Wang et al.~\cite{POAC} demonstrate the potential of MORL for end-to-end AD by learning Pareto-optimal policies over objectives like speed and fuel efficiency using image-based input. However, the approach optimizes only a single preference configuration per training run, requiring retraining for new preferences and thus limiting runtime adaptability.

Recent single-policy MORL frameworks show promise for runtime adaptability of preferences~\cite{PD-MORL, JorgeNew}, but have not yet been applied to vision-based end-to-end autonomous driving. 
We extend that line by integrating an adaptable, preference driven framework into high-dimensional, vision-based urban driving scenarios, as illustrated in Figure~\ref{fig:approach}. 

\section{Methodology}
\subsection{Preference-Based Driving Task}
We consider an RL-driven autonomous vehicle navigating through dense urban environments with dynamic traffic under varying weather and lighting conditions, trained and evaluated with the CARLA simulator \cite{Carla}.
The agent’s objective is to navigate a predefined route represented by waypoints, while avoiding collisions and adapting its behavior to given human driving preferences. 
These preferences—aggressiveness, comfort, speed and efficiency—may vary with time or task and are encoded as a four-dimensional vector to modulate the agent's behavior.
The agent receives semantic segmentation (SEG) from a front-facing camera as input and has access to the vehicle’s odometry, the current speed limit, and the next waypoint. While it must comply with the current speed limit, advanced traffic rules are not considered.
The vehicle is controlled via continuous steering and throttle/brake commands as direct output from the agent. 

\subsection{Multi-Objective Reinforcement Learning}
MORL extends traditional reinforcement learning by optimizing multiple, often conflicting objectives~\cite{hayes}. 
The agent is tasked with learning policies that reflect trade-offs across these objectives by utilizing a vector-valued reward function.
Formally, the learning problem is defined as a Multi-Objective Markov Decision Process (MO-MDP), specified by the tuple $(\mathcal{S}, \mathcal{A}, P, \mathcal{R}, \gamma)$, where $\mathcal{S}$ denotes the state space, $\mathcal{A}$ the action space, $P: \mathcal{S} \times \mathcal{A} \times \mathcal{S} \rightarrow [0,1]$ the transition dynamics, and $\gamma \in [0,1]$ the discount factor. A key component is the vector-valued reward function $\mathcal{R}: \mathcal{S} \times \mathcal{A} \rightarrow \mathbb{R}^n$, which returns $n$ objective-specific rewards at each time step.
A single policy $\pi^{\boldsymbol{\lambda}}$ can be conditioned on a user-defined preference vector $\boldsymbol{\lambda} \in \mathbb{R}^n$ to express trade-offs between multiple objectives.
The preferences are incorporated into the learning process via element-wise multiplication with the Q-values  \(\boldsymbol{\lambda} \odot \mathbf{Q}_j \).

We employ the preference-driven (PD-)MORL TD3 implementation by Basaklar~\etal~\cite{PD-MORL}, specifically their MO-TD3-HER variant, and adapt it to the domain of end-to-end autonomous driving. 
PD-MORL trains a single policy to generalize across the entire preference space by extending the standard TD3 actor-critic framework through two core mechanisms:
(i) During policy optimization, the alignment between user preferences and the critic’s learned Q-values is improved by projecting the original preference vector $\boldsymbol{\lambda}$ into a normalized latent space using a preference interpolator $\mathcal{I}(\boldsymbol{\lambda}) = \boldsymbol{\lambda}_p$. This allows augmenting the loss function with an angle loss term $\mathcal{L}_{\text{angle}}(\boldsymbol{\lambda}_p, \mathbf{Q})$, which penalizes directional mismatches between the interpolated preference vector $\boldsymbol{\lambda}_p$ and the critic’s output $\mathbf{Q}$.
(ii) A preference-aware hindsight experience replay (HER) mechanism enhances generalization across the entire preference space by resampling experiences using alternative preference vectors.

While PD-MORL \cite{PD-MORL} was originally evaluated on low-dimensional gym benchmarks~\cite{benchmark-MORL}, we extend the framework to a high-dimensional, vision-based autonomous driving task in dense, mixed-traffic environments. 
Our focus lies on preference-aware behavioral adaptation and safe navigation.
To the best of our knowledge, this is the first application of PD-MORL in a vision-based autonomous driving context that supports runtime preference adaptation. 

\subsection{State and Action Space}
The agent's state $\boldsymbol{s}$ consists of visual perception, vehicle measurements, navigation context, and user preferences.

As primary input, the agent receives segmentation images from a front-facing camera with a resolution of $448 \times 224$ pixels and a field of view of $110^\circ$. 
The wider horizontal field of view reflects the increased relevance of lateral information in AD.
The images are symmetrically zero-padded along the vertical axis and resized to $224 \times 224$ before being processed by a pretrained ResNet18-based feature extractor, resulting in the feature vector $\boldsymbol{f} \in \mathbb{R}^{128}$.

Vehicle odometry $\boldsymbol{o}$ includes velocity and acceleration (both magnitudes and directional components), along with the current positions of the throttle, brake, and steering controls.
By incorporating the previous three control inputs $\boldsymbol{a}_{\text{hist}}$ into the state, temporal dependencies in vehicle behavior are captured, assisting the agent in learning smooth vehicle dynamics \cite{smoother, osci, Transfuser}.

For navigation, the agent receives a polar coordinate representation of the next waypoint $\boldsymbol{wp}$, including the Euclidean distance and the angular offset $\theta_{\text{nav}} \in [-\pi, \pi]$, as well as the current speed limit $v_{\text{limit}}$.

The final state space can be expressed as:
\[
\boldsymbol{s}  = (\boldsymbol{f}, \boldsymbol{o}, \boldsymbol{a}_{\text{hist}}, \boldsymbol{wp}, v_{\text{limit}})
\] 

The action space is continuous and two-dimensional, consisting of a steering adjustment $\theta \in [-1, 1]$ and a throttle/brake signal $u \in [-1, 1]$. In CARLA, throttle and brake commands are separately scaled within $[0,1]$: values $u > -0.5$ are mapped linearly to throttle intensity, whereas $u < -0.5$ are mapped to braking intensity. The steering adjustment $\theta$ modifies the vehicle's heading by up to $\pm11^\circ$ per timestep. The final continuous action vector is defined as $\boldsymbol{a} = (\theta, u)$.
The state-action loop is running at $10\,\mathrm{Hz}$.

\subsection{Networks}
The agent's learning architecture consists of multi-layer perceptron networks with ReLU activations for both the actor and the two critics, using the architecture $[\text{input} \rightarrow 250 \rightarrow 125 \rightarrow \text{output}]$. 
The input consists of the state $\boldsymbol{s} \in \mathbb{R}^{154}$, the preference vector $\boldsymbol{\lambda} \in \mathbb{R}^{4}$, and the action $\boldsymbol{a} \in \mathbb{R}^{2}$ (for the critic only). 
The output is the action $\boldsymbol{a}$ for the actor and a Q-value vector $\boldsymbol{Q} \in \mathbb{R}^{5}$ for the critic.

A shared visual feature extractor is prepended to both the actor and the critics. The encoder is implemented as a truncated ResNet-18 with six residual blocks and pretrained ImageNet weights~\cite{ResNet}. The feature extractor's weights are initially frozen for the first 7{,}200 gradient steps to promote Q-value stability, while the total training comprises 1{,}000{,}000 steps.

\subsection{Training}
Training is conducted on random routes across four distinct CARLA towns (Town01–Town04) with varying weather and lighting. For quantitative evaluation under structured and reproducible conditions, we define seven fixed evaluation scenarios, each spanning 300–800 meters and distributed across the four towns. The scenarios incorporate diverse elements such as roundabouts, tunnels, highways, junctions, elevation changes, and varying traffic densities.

Each episode begins at a random location and consists of 500 waypoints spaced approximately 5 meters apart. 
Termination occurs upon collisions, deviation from the route by more than 6 meters, stagnation (no progress for 200 steps), successful completion of the route, or after a maximum of 1,400 environment steps (approximately 140 seconds).
Simulated traffic is handled by CARLA, following realistic driving behavior and complying with speed limits. However, traffic light adherence is beyond the scope of this work and is therefore ignored.

\subsection{Reward Vector}\label{sec:Reward_Vector}
The reward vector $\boldsymbol{r_t}$ captures both the core driving task and user-dependent driving style objectives. 
The core driving reward occupies the first entry in the vector and captures essential driving and navigation competence.
The remaining four components correspond to distinct driving behavioral preferences: aggressiveness, comfort, speed and efficiency.

At each timestep $t$, the agent receives a five-dimensional reward vector:

\[
\boldsymbol{r_t} = (
\underbrace{r_t^{\text{core}}}_{\text{static}},\ 
\underbrace{r_t^{\text{agg}},\ r_t^{\text{comfort}},\ r_t^{\text{speed}},\ r_t^{\text{eff}}}_{\text{preference-based (dynamic) - }\boldsymbol{r_{\text{pref}}}}).
\]

The preference-based objectives are weighted by a user-defined continuous preference vector $\boldsymbol{\lambda} \in [0,1]^4$, supporting smooth transitions and combinations of different driving styles. 
During training, $\boldsymbol{\lambda}$ is sampled randomly at the beginning of each episode (normalized and convex); during evaluation, it is given to reflect a target driving style. 

\begin{figure}[t]
	\centering
	\includegraphics[width=\linewidth]{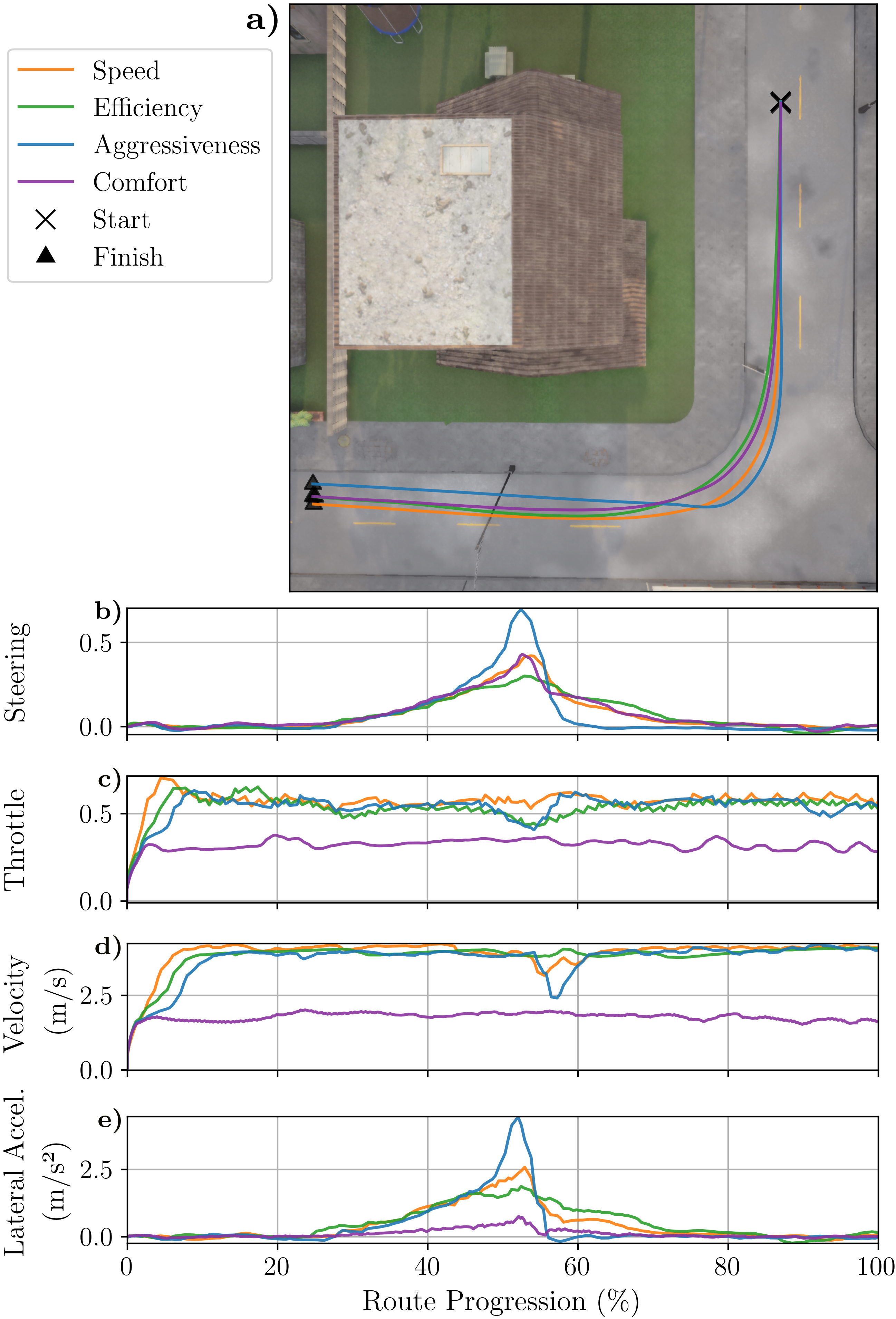}
    	\caption[Trajectories and Dynamics for Individual Preference Objectives \MakeUppercase{\romannumeral 1}]{\textbf{a)} Bird’s-eye view of the agent’s trajectories while navigating a T-intersection. Each trajectory reflects a distinct driving style, achieved by setting one preference weight (for \textcolor{SpeedColor}{speed}, \textcolor{EfficiencyColor}{efficiency}, \textcolor{AggressivenessColor}{aggressiveness}, or \textcolor{ComfortColor}{comfort}) to one, with all others set to zero. Subfigures \textbf{b)} - \textbf{e)} display additional metrics including steering, throttle, velocity, and lateral acceleration. Values are smoothed using exponential smoothing with a factor of $\beta = 0.6$. }
	\label{fig:corner}
    \vspace*{0.4em}
\end{figure}

\subsubsection{Core Driving Reward}
The static core reward remains unaffected by user preferences and captures fundamental driving skills. It encourages route progression, braking in hazardous situations, and maintaining an appropriate velocity, while penalizing unsafe behaviors, including collisions, off-road driving, lane invasions, and speed limit violations~\cite{baselineRWD, baselineWp}. The core 
reward is composed of five components:

\[
r_{\text{core}} = r_{\text{collision}} + r_{\text{boundary}} + r_{\text{lane}} + r_{\text{nav}} + r_{\text{perf}}.
\]

The detailed formulation of each component is provided in the Appendix~\ref{core_driving_reward}. Combined, these elements ensure stable and safe driving behavior regardless of user preference modulation.

\subsubsection{Preference-Based Objectives}
The preference-based reward vector $\boldsymbol{r}_{\text{pref}}$ encodes driving style objectives, with each dimension corresponding to a specific preference, guided by concepts from \cite{POAC, baselineRWD, speed}.

\textbf{Aggressiveness} ($r_{\text{agg}}$): Encourages a sporty driving style by rewarding high accelerations and yaw rates, while penalizing unstable longitudinal behavior (to avoid brake-gas oscillation). Computed as:
    \[
    r_\text{agg} = \alpha_{\text{l}} \cdot  |a_{\text{long}}| + \alpha_{\text{l}} \cdot  |a_{\text{lat}}| + \alpha_{\text{yaw}} \cdot  |r_{\text{yaw}}| - \alpha_{\text{l\_acc}} \cdot  |\Delta a_{\text{long}}|,
    \]
    where rapid sign changes in longitudinal acceleration (unstable brake-gas transitions) trigger the penalty term.

\textbf{Comfort} ($r_t^{\text{comfort}}$): Promotes smooth driving by penalizing abrupt changes in velocity, acceleration, control signals (steering, throttle), and the magnitude of jerk. Computed as:
     \[
    \begin{aligned}
    r_\text{comfort} =\ & - \beta_{\text{steer}}  \cdot |\Delta \theta|
    - \beta_{\text{throttle}} \cdot |\Delta u| - \beta_{\text{v}} \\
     & \cdot |\Delta v| 
    - \beta_{\text{long}} \cdot |\Delta a_{\text{long}}| 
    - \beta_{\text{jerk}} \cdot \|\text{jerk}\| + \beta_{\text{b}}.   
    \end{aligned}
    \]
    
\textbf{Speed} ($r_{\text{speed}}$): Rewards matching the target velocity $v_{\text{target}}$ based on the speed limit, with deviations from the target velocity penalized as:
    \[
    r_\text{speed} = -\delta_{\text{speed}} \cdot \left( \frac{|v_{\text{target}} - v_{\text{current}}|}{v_{\text{target}}} \right) + \delta_{\text{speed}} .
    \]    
     
\textbf{Efficiency} ($r_{\text{eff}}$): Encourages fuel-efficient driving by minimizing throttle as well as acceleration, and promoting moderate speeds. It is approximated by:
    \[
    r_\text{eff} = (1 - u) \cdot \frac{v_{\text{current}}}{v_{\text{max}}} \cdot \left(1 - \frac{a_{\text{current}}}{a_{\text{max}}}\right) .
    \]

All parameters are provided in Table~\ref{tab:weights}.

\begin{figure*}[t]
	\centering
	\includegraphics[width=0.99\linewidth]{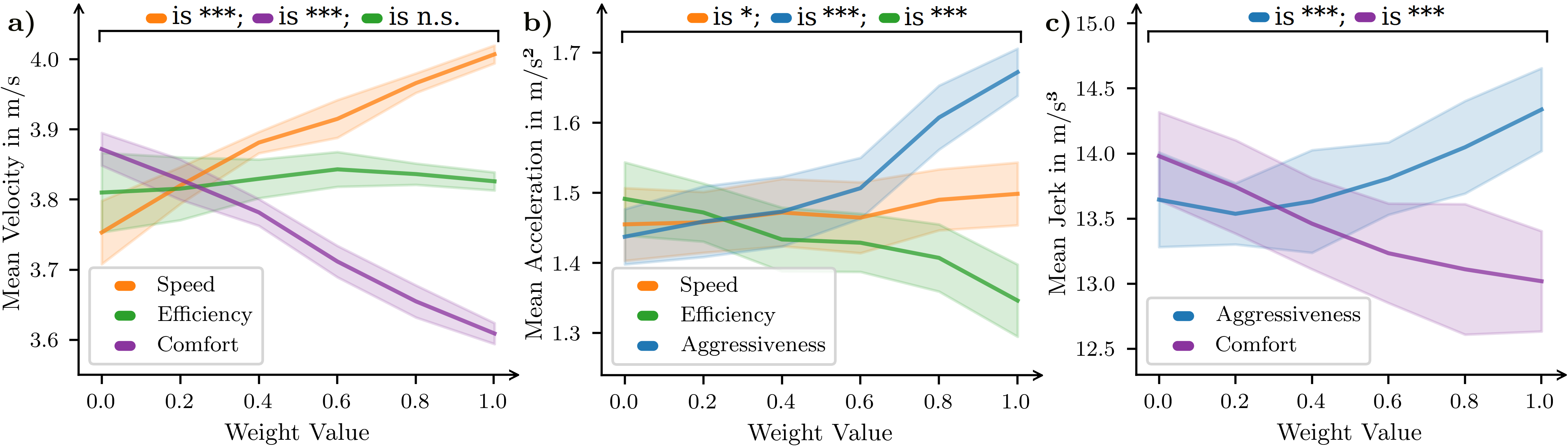}   
    \caption[Effect of Preference Weights on Driving Behavior]{
    Effect of increasing preference weights on key driving behavior metrics. Each subplot illustrates the relationship between a specific driving behavior metric (velocity, acceleration, or jerk) and the assigned weight of an individual preference objective. The shared x-axis denotes the fixed sampled weight for the respective objective, while the y-axis shows the corresponding mean value observed per episode. 
    For each weight level, the metric is observed in 20 episodes, with the remaining preference weights randomly sampled under the convexity constraint. Solid lines indicate the mean trend, and shaded regions represent one standard deviation.
    Statistical significance between weight settings ($0$ vs.~$1$) is assessed using Welch’s t-tests, with standard significance levels indicated by $\mathrm{n.s.}$ ($p \geq 0.05$), $^*$ ($p < 0.05$), $^{**}$ ($p < 0.01$), and $^{***}$ ($p < 0.001$).
    \textbf{a)}~Mean velocity increases with higher \textcolor{SpeedColor}{speed} weighting and decreases under \textcolor{ComfortColor}{comfort}, consistent with their respective behavioral emphasis, while remaining stable for \textcolor{EfficiencyColor}{efficiency}. 
    \textbf{b)}~Mean acceleration increases with \textcolor{AggressivenessColor}{aggressiveness} weighting and slightly decreases under \textcolor{EfficiencyColor}{efficiency}, in line with the intended driving style effects.   
    \textbf{c)}~Mean jerk decreases with increased \textcolor{ComfortColor}{comfort} preference and increases under \textcolor{AggressivenessColor}{aggressiveness}, indicating smoother or more dynamic driving styles, respectively.
}
	\label{fig:prefs}
\end{figure*}

\vspace*{0.1em}
\subsection{Driving and Preference Metrics}\label{subsec:metricies}
\vspace*{0.1em}
To assess both driving performance and preference adherence, we evaluate the following set of metrics. \textbf{Route Completion} ($RC$), \textbf{Episode Duration} ($T$), and \textbf{Episode Reward} ($R$) reflect core task performance in terms of spatial coverage, runtime, and cumulative reward, respectively.

The \textbf{Driving Score}, as defined by the CARLA Leaderboard~\cite{CARLA_Leaderboard}, is given by $\textit{DS} = RC \times \prod_i p_i^{n_i}$ and summarizes overall driving quality by combining route completion with a penalty term for infractions, where $p_i$ is the penalty factor and $n_i$ the number of occurrences of infraction type~$i$, including collisions, timeouts, speeding, and lane violations.

Safety-related metrics include the vehicle and environment \textbf{Collision Rate}, defined as $\textit{CR} = N_{\text{collisions}} / T$, as well as the \textbf{Lane Invasion Rate} $\textit{LIR} = N_{\text{invasions}} / T$, which measures lane boundary crossings into oncoming traffic lanes. The \textbf{Lane Deviation} $\textit{LD}$ reflects the average lateral offset from the road centerline and is given by $\textit{LD} = \frac{1}{T} \sum_{t=1}^T d_{\text{lat},t}$.

To assess preference adherence, we report the \textbf{Preference Score}, defined as $\textit{PS} = \frac{1}{T} \sum_{t=1}^{T} \boldsymbol{\omega}^\top \boldsymbol{r}_{\text{Prefs}, t}$. It is calculated as the preference-weighted return over the episode, where $\boldsymbol{\omega}$ is the user-defined preference vector and $\boldsymbol{r}_{\text{Prefs}, t}$ the vector of preference-specific rewards at time~$t$.

Finally, \textbf{Preference Alignment} quantifies how well the critic’s Q-values reflect the intended preferences, computed as $\textit{PA} = \frac{1}{N} \sum_{t=1}^{N} g(\boldsymbol{\omega}_t, \boldsymbol{Q}(\boldsymbol{s}_t, \boldsymbol{a}_t, \boldsymbol{\omega}_t))$, where $g(\cdot)$ is the cosine-based angular deviation (in degrees) between preference vector and Q-values. Lower \textit{PA} values indicate stronger alignment and serve as a quantitative measure of how faithfully the critic encodes user-specified preferences.

\section{Experimental Evaluation}\label{sec:exp}
To assess the effectiveness of our proposed approach, we conduct a series of experiments designed to evaluate both driving performance and the ability to modulate behavior according to user preferences. The evaluation is guided by the following research questions:

\begin{rqenum}
    \item Can multi-objective reinforcement learning effectively integrate user-defined preferences into vision-based end-to-end autonomous driving?
    \item How do different feature extractors and input modalities affect driving performance and preference alignment in complex urban scenarios?
\end{rqenum}

\begin{table*}	
	\centering
	{\footnotesize
        \begin{tabularx}{\textwidth}{|X|Y|Y|Y|Y|Y|Y|Y|}
			\hline
			\rowcolor{white}
			& \textbf{Ours} & \multicolumn{2}{c|}{\textbf{Feature Extractor Ablation}} & \multicolumn{4}{c|}{\textbf{Vision Modality Ablation}} \\
			\rowcolor{white}
            \hline
			\textbf{Metric} & ResNet~[SEG] & MobileNet & CNN & RGB & Depth & RGB-D & SEG-D \\

			\hline
			\rowcolor{lightgray}
			Driving Score ($\%$) $\uparrow$  & \textbf{42 $\pm$ 17} & 35 $\pm$ 14 & 19 $\pm$ 7 & 23 $\pm$ 8 & 28 $\pm$ 9 & 25 $\pm$ 10 & 31 $\pm$ 12 \\
			\rowcolor{darkgray}
			Prefer. Score (cont.) $\cdot 10^2$~$\uparrow$ & \textbf{38 $\pm$ 3} & 35 $\pm$ 4 & 35 $\pm$ 2 & 34 $\pm$ 2 & 34 $\pm$ 2 & 35 $\pm$ 3 & 34 $\pm$ 4 \\
            \hline
			\rowcolor{lightgray}
			Preference Alignment ($\circ$) $\downarrow$ & \textbf{11.30 $\pm$ 0.06} & 19.15 $\pm$ 0.46 & 11.33 $\pm$ 0.04 & 13.29 $\pm$ 0.25 & 14.06 $\pm$ 0.14 & 15.62 $\pm$ 0.35 & 12.41 $\pm$ 0.08 \\
			\rowcolor{darkgray}
			Route Completion ($\%$) $\uparrow$ & \textbf{57 $\pm$ 16} & 52 $\pm$ 13 & 33 $\pm$ 10 & 40 $\pm$ 10 & 45 $\pm$ 11 & 43 $\pm$ 11 & 48 $\pm$ 14 \\
			\rowcolor{lightgray}
			Vehi. Collision Rate $\cdot 10^2\downarrow$ & \textbf{0.40 $\pm$ 0.43} & 0.87 $\pm$ 0.41 & 0.61 $\pm$ 0.46 & 0.59 $\pm$ 0.25 & 0.71 $\pm$ 0.30 & 0.90 $\pm$ 0.59 & 1.02 $\pm$ 0.32 \\
			\rowcolor{darkgray}
			Env. Collision Rate $\cdot 10^2$~$\downarrow$ & 0.38 $\pm$ 0.38 & \textbf{0.19 $\pm$ 0.18} & 0.37 $\pm$ 0.35 & 0.62 $\pm$ 0.66 & 0.47 $\pm$ 0.56 & 0.41 $\pm$ 0.31 & 0.24 $\pm$ 0.60 \\
			\rowcolor{lightgray}
            Lane Invasion Rate $\cdot 10$~$\downarrow$& 0.32 $\pm$ 0.05 & 0.34 $\pm$ 0.13 & 0.37 $\pm$ 0.21 & 0.34 $\pm$ 0.15 & 0.30 $\pm$ 0.14 & \textbf{0.26 $\pm$ 0.13} & 0.29 $\pm$ 0.10 \\
			\rowcolor{darkgray}    
			Lane Deviation (m) $\downarrow$ & 1.31 $\pm$ 0.34 & 1.23 $\pm$ 0.20 & 1.13 $\pm$ 0.12 & 1.26 $\pm$ 0.22 & 1.20 $\pm$ 0.20 & 1.40 $\pm$ 0.30 & \textbf{1.09 $\pm$ 0.30} \\
			\hline
            
		\end{tabularx}
	}
    \caption[Performance Comparison Across Architectures and Modalities]{
    Quantitative performance comparison of feature extractors (MobileNet, CNN, ResNet) and input modalities (RGB, Depth, SEG, RGB-D, SEG-D). 
    Ours refers to the best-performing configuration, which leverages segmentation-based input and a modified ResNet encoder built upon a truncated ResNet-18 architecture comprising six residual blocks.
    Each configuration was evaluated over ten agent seeds and seven diverse driving scenarios in CARLA, using 540 sampled preference vectors per agent to cover the full preference space. 
    Metrics are reported as mean~$\pm$~standard deviation, averaged across all scenarios and preferences.
    }
	\label{tab:Merged}
\end{table*}

\subsection{Qualitative Preference Analysis}
Figure~\ref{fig:corner} shows a qualitative analysis of the agent’s driving behavior under different preference vectors, presenting both top-down view trajectories and corresponding driving metric graphs.

The scenario involves navigating from a fixed start point to a goal location, making a right turn at a T-intersection.
The bird’s-eye view displays the agent's trajectories under four different preference settings, as shown in Fig.~\ref{fig:corner}a. Each trajectory is obtained by setting one preference weight to $1.0$ while keeping all others at zero, enabling a comparison of individual objectives in their most pronounced form. The start and end points are marked by "X" and \(\blacktriangle\), respectively.

Fig.~\ref{fig:corner}b-e display key behavioral metrics along the route: steering angle (b), throttle (c), velocity (d), and longitudinal acceleration (e). All metrics are plotted over route progression, with each x-axis step corresponding to one meter along the optimal path. Values are smoothed using exponential smoothing with factor $\beta = 0.6$.

While all trajectories successfully reach the goal and remain within lane boundaries (a), distinct preference-driven behaviors become evident through the driving metrics.
The \textcolor{SpeedColor}{speed}-focused policy reaches higher velocities early on—peaking at approximately 4.1~m/s—as shown in the velocity plot (d). In contrast, the \textcolor{ComfortColor}{comfort}-oriented policy maintains a consistently lower and smoother velocity profile, averaging around 2~m/s. It also exhibits smooth cornering with a low lateral acceleration of 0.9~m/s² (e), indicating a gentle and stable driving style. Conversely, the \textcolor{AggressivenessColor}{aggressiveness} policy reaches a peak lateral acceleration of 4.2~m/s² (e) and displays sharper steering angles, characteristic of a more dynamic and sporty behavior, while exhibiting a drop in velocity during cornering (d). Finally, the \textcolor{EfficiencyColor}{efficiency} policy shows gradual steering (b) and reduced throttle input (c), yet maintains a high velocity (d), reflecting energy-efficient driving behavior.

The results show that the policy adapts to different preference configurations, displaying distinct and interpretable driving styles.
A supplemental video illustrating the agent’s qualitative behavior is available online.\footnote{Supplemental video: \url{https://youtu.be/2brpyC_edHw}}

\subsection{Quantitative Analysis}
\subsubsection{Preference Reflection}
We conduct a quantitative evaluation of preference reflection by analyzing its impact on key driving metrics—velocity, acceleration, and jerk—in an urban driving scenario, as shown in Figure~\ref{fig:prefs}.

To illustrate the influence of each objective on behavior, every preference objective is systematically assessed by stepwise varying its weight in the interval $[0,1]$, while the remaining preferences are randomly sampled under the convexity constraint. The total evaluation spans 480 episodes ($20$ episodes per weight level, $6$ levels, $4$ objectives).

Each subplot shows a specific driving metric relevant to one or more driving objectives, with mean (solid line) and standard deviation (shaded area) calculated across evaluation episodes.
The selected preferences in each subplot correspond to their conceptually matching driving metric. The x-axis represents the fixed sampled weight of the respective objective, shared across all subplots.

Also, we statistically assess the policy's sensitivity to preference changes reflected in the driving metrics above, using Welch’s t-tests for difference in distribution between extreme weight configurations ($0$ vs.~$1$) for each objective-metric pair. 

Mean velocity significantly increases with higher speed weighting and decreases under comfort (both $p < 0.001$), reflecting their respective emphasis on fast versus smooth driving, as shown in Fig.~\ref{fig:prefs}a. 
In contrast, velocity remains largely constant under efficiency (no significant change), indicating that the agent does not trivially conserve energy by slowing, but instead maintains a steady speed to optimize throttle usage.

Mean acceleration magnitude increases significantly with higher aggressiveness weighting ($p < 0.001$), representing a more dynamic and responsive driving style, as shown in Fig.~\ref{fig:prefs}b. 
Conversely, it decreases for efficiency ($p < 0.001$), consistent with smoother acceleration profiles for energy savings. 
A change in the speed objective yields less change in acceleration values, as the agent learns to maintain high speed through early acceleration followed by cruising.

Stronger comfort preferences lead to significantly lower mean jerk magnitude ($p < 0.001$), indicating smoother control, while higher aggressiveness preferences increase jerk (also $p < 0.001$), resulting in more abrupt driving behavior, as shown in Fig.~\ref{fig:prefs}c.

In support of RQ1, the results confirm that varying the preference weights induces significant shifts in driving behavior, demonstrating the policy's responsiveness to various preferences. 

\subsubsection{Driving and Preference Performance}
Beyond preference reflection, we assess overall driving and preference performance to identify the most effective model configuration. This evaluation includes a systematic comparison of different visual encoders and input modalities across key metrics (see Table~\ref{tab:Merged}), covering both task-specific driving performance and preference-related indicators (see Section~\ref{subsec:metricies}).

Each configuration is evaluated over seven diverse driving scenarios in CARLA, using ten independently trained and randomly seeded agents. For each agent, 540 unique user preference vectors are sampled using a dense strategy that ensures full coverage of the preference space. Final results are reported as mean~$\pm$~standard deviation, aggregated across all agents, scenarios, and preference combinations.

\paragraph{Feature Extractor Ablation}
To analyze the impact of visual representation quality, we ablate our well-performing truncated 6-block ResNet18 variant~\cite{5blockresnet} against a lightweight CNN (NatureCNN~\cite{NatureCnnDQN}) and an efficient MobileNet~\cite{MobileNetv3}, evaluating differences in capacity and architecture. ResNet and MobileNet use pretrained ImageNet weights, whereas the CNN is trained from scratch. All models are adapted from prior work to reflect distinct trade-offs between computational efficiency, model capacity, and representational power.

Our configuration, which combines semantic segmentation input with the ResNet-based encoder, achieves the best overall performance across all categories. It yields the highest driving score ($42 \pm 17$), preference score ($38 \pm 3$), and route completion rate ($57 \pm 16\%$), while maintaining low vehicle collision and lane invasion rates. Crucially, it also achieves the lowest preference alignment angle ($11.3^\circ$), indicating the strongest consistency between the learned policy and user-defined preferences. 
While MobileNet achieved a lower environment collision rate ($0.19 \pm 0.18$), it exhibited a worse preference alignment angle ($19.15 \pm 0.46^\circ$) and a lower preference score ($35 \pm 4$), suggesting a more conservative, yet less preference-aligned driving behavior.

Across all visual encoders, our ResNet-based model consistently outperforms both CNN and MobileNet variants, emphasizing the importance of feature expressiveness and pretraining for learning effective preference-aware policies.

\paragraph{Vision Modality Ablation}
We further ablate the effect of different visual input modalities on driving and preference performance. While RGB-D and SEG-D modalities show improvements in specific sub-metrics (e.g., lane deviation or lane invasion), they remain inferior in driving and preference scores compared to the pure SEG-based configuration. This suggests that added input complexity may introduce distractive information or shift attention away from preference-aligned behavior.

Across all modalities, SEG input paired with a ResNet encoder performs best, highlighting the importance of focused semantic representations and high-quality visual encoding. We find that both the choice of feature extractor and input modality significantly affect driving performance and preference alignment, thereby addressing RQ2.

Taken together, our experimental results affirm the framework’s capability to drive in urban environments while dynamically adapting to user preferences and successfully exhibiting personalized driving behaviors.

\section{Conclusion}\label{sec:conclusion}
In summary, we present a multi-objective reinforcement learning approach capable of integrating dynamic user preferences into end-to-end vision-based autonomous driving in the CARLA simulator.
Our method employs a single policy conditioned on a continuous, user-defined preference vector, enabling runtime adaptation of driving behavior without retraining. Leveraging vision-based perception, the agent learns to modulate its driving style in response to varying preference weightings.
Experimental evaluations in complex urban traffic scenarios demonstrate that the agent maintains robust driving performance while reflecting user preferences across diverse metrics. 
To improve perception quality, we evaluate multiple visual input modalities and feature extraction architectures, with segmentation input and a compact ResNet-based encoder achieving the best overall performance.
Future work could extend this framework by modeling more sophisticated traffic rules and validating it on real-world driving platforms.

\bibliographystyle{IEEEtran}
\bibliography{surmann_bib}
\vspace{0.1em}
\appendix

\subsection{Core Driving Reward Components}\label{core_driving_reward} 
The core driving reward is composed of five components that encourage safe and effective navigation.
The collision penalty term discourages collisions with vehicles and obstacles by applying a scaled penalty based on impact acceleration, while rewarding proactive braking when hazards are detected:\newline
$r_{\text{collision}} = - w_{\text{type}} (c_{\text{col}} + c_{\text{acc}} \cdot a) \cdot \mathbb{1}_{\text{col}} + c_{\text{brake}} \cdot \mathbb{1}_{\text{brake}}$

Boundary control penalizes off-road driving and lane invasions, weighted by the severity of the infraction:\newline
$r_{\text{boundary}} = - c_{\text{off}} \cdot \mathbb{1}_{\text{off-road}} - w_{\text{lane}} c_{\text{inv}} \cdot \mathbb{1}_{\text{invasion}}$

Lane deviation penalizes lateral offset from the center of the lane to encourage stable lane-keeping:\newline 
$r_{\text{lane}} = -c_{\text{dev}} \cdot d_{\text{center}}$

Navigation rewards progress along waypoints while penalizing lateral deviation from the planned route and heading misalignment:\newline
$r_{\text{nav}} = c_{\text{wp}} + P_{\text{prog}} - c_{\text{lat}} d_{\text{lat}} - c_{\text{head}} |\theta_{\text{head}}| - d_{\text{route}}$

Finally, performance regularization penalizes unsafe or inefficient driving actions, depending on current situation:\newline
$
r_{\text{perf}} = 
\begin{cases}
r_{\text{speed}}, & \text{if clear path} \\
- c_{\text{spd\_high}} |v - v_{\text{limit}}|, & \text{if } v > v_{\text{limit}} \\
- c_{\text{idle}}, & \text{if idle} \\
- c_{\text{osc}}, & \text{if oscillating} \\
- c_{\text{steer}} |\Delta \delta| - c_{\text{throttle}} |\Delta \tau|, & \text{if abrupt actions} \\
\end{cases}
$

All coefficients \( c_* \), distance and angle terms, and indicators are defined in Table~\ref{tab:weights}. 

\begin{table}[!t]
\centering
\begin{tabularx}{\linewidth}{llX}
    \textbf{Notation} & \textbf{Value} & \textbf{Description} \\
    \hline
    \multicolumn{3}{l}{\textit{Preference Reward Parameters:}} \\
    \(\alpha_{\text{l}}\) & 0.10 & Weight for lat. and longit. accel. magnitude \\
    \(\alpha_{\text{yaw}}\) & 0.30 & Weight for yaw rate (Aggressiveness) \\
    \(\alpha_{\text{l\_acc}}\) & 0.20 & Penalty for longitudinal acceleration change \\
    \(\beta_{\text{steer}}\) & 0.10 & Penalty for abrupt steering changes \\
    \(\beta_{\text{throttle}}\) & 0.05 & Penalty for throttle variability  \\
    \(\beta_{\text{v}}\) & 0.30 & Penalty for abrupt velocity changes  \\
    \(\beta_{\text{long}}\) & 0.30 & Penalty for longitudinal acceleration change \\
    \(\beta_{\text{jerk}}\) & 0.03 & Penalty for jerk magnitude  \\
    \(\beta_{\text{b}}\) & 1.20 & Constant bias added to comfort reward \\
    \(\delta_{\text{speed}}\) & 1.75 & Weight for deviation from target velocity \\
    \hline
    \multicolumn{3}{l}{\textit{Core Reward Parameters:}} \\
    \(c_{\text{col}}\) & 5 & Base penalty for collisions \\
    \(c_{\text{acc}}\) & 0.1 & Acceleration-based penalty in collisions \\
    \(c_{\text{brake}}\) & 2.75 & Reward for braking when obstacle ahead \\
    \(w_{\text{type}}\) & 1.7 & Collision type weight (env. > vehicle) \\

    \(c_{\text{spd\_high}}\) & 0.3 & Penalty for exceeding speed limit \\
    \(c_{\text{idle}}\) & 3.5 & Penalty for unnecessary idling \\
    \(c_{\text{osc}}\) & 0.2 & Penalty for oscillatory steering \\
    \(c_{\text{steer}}\) & 0.6 & Penalty for fast steering changes \\
    \(c_{\text{throttle}}\) & 0.4 & Penalty for abrupt throttle changes \\
    
    \(c_{\text{off}}\) & 1.2 & Penalty for off-road driving \\
    \(c_{\text{inv}}\) & 1 & Penalty for lane invasion \\
    \(w_{\text{lane}}\) & 2 & Lane invasion severity weight \\
    
    \(c_{\text{dev}}\) & 0.15 & Penalty for lateral lane deviation \\
    
    \(c_{\text{lat}}\) & 1/3 & Penalty for lateral route deviation \\
    \(c_{\text{head}}\) & 1/90 & Penalty for heading misalignment \\

    \(P_{\text{prog}} \) & – & Relative progress between waypoints \\
    \(c_{\text{wp}}\) & 0.25 & Reward for reaching waypoints, inclduding: \\
    \( c_{\text{junc}}\) & 0.1 & Bonus for junction traversal \\
    \( c_{\text{goal}}\) & 20 & Bonus for reaching the final goal \\
    \( c_{\text{termination}}\) & -5 & Penalty for episode termination \\
    \(\mathbb{1}_{\text{condition}}\) & -- & Indicator function, evaluates if the event occurs \\ 
    \hline
\end{tabularx}
\caption{Weight parameters for preference-based objectives and core reward components.}
\label{tab:weights}
\end{table}

\end{document}